\documentclass[a4paper, 10pt, conference]{IEEEtran}      
\IEEEoverridecommandlockouts                              
\IEEEpubid{\makebox[\columnwidth]{\textcopyright 2020 IEEE. Accepted at Intelligent Vehicles Symposium (IV) 2020.\hfill} \hspace{\columnsep}\makebox[\columnwidth]{ }}

\usepackage{graphics} 
\usepackage{epsfig} 
\usepackage{mathptmx} 
\usepackage{times} 
\usepackage{amsmath} 
\usepackage{amssymb}  

\usepackage{gensymb}  
\usepackage{subcaption}  
\usepackage[ruled,vlined]{algorithm2e}  
\usepackage{booktabs}  
\usepackage{makecell}  
\usepackage{colortbl}  
\usepackage{multirow}  
\usepackage{capt-of} 
\usepackage{hyperref} 

\usepackage{pgfplots}  
\usepgfplotslibrary{units}  

\graphicspath{{./figures/}}

\definecolor{unlabeled}{rgb}{0.0, 0.0, 0.0}
\definecolor{car}{rgb}{0.39215686274509803, 0.5882352941176471, 0.9607843137254902}
\definecolor{bicycle}{rgb}{0.39215686274509803, 0.9019607843137255, 0.9607843137254902}
\definecolor{motorcycle}{rgb}{0.11764705882352941, 0.23529411764705882, 0.5882352941176471}
\definecolor{truck}{rgb}{0.3137254901960784, 0.11764705882352941, 0.7058823529411765}
\definecolor{othervehicle}{rgb}{0.0, 0.0, 1.0}
\definecolor{person}{rgb}{1.0, 0.11764705882352941, 0.11764705882352941}
\definecolor{bicyclist}{rgb}{1.0, 0.1568627450980392, 0.7843137254901961}
\definecolor{motorcyclist}{rgb}{0.5882352941176471, 0.11764705882352941, 0.35294117647058826}
\definecolor{road}{rgb}{1.0, 0.0, 1.0}
\definecolor{parking}{rgb}{1.0, 0.5882352941176471, 1.0}
\definecolor{sidewalk}{rgb}{0.29411764705882354, 0.0, 0.29411764705882354}
\definecolor{otherground}{rgb}{0.6862745098039216, 0.0, 0.29411764705882354}
\definecolor{building}{rgb}{1.0, 0.7843137254901961, 0.0}
\definecolor{fence}{rgb}{1.0, 0.47058823529411764, 0.19607843137254902}
\definecolor{vegetation}{rgb}{0.0, 0.6862745098039216, 0.0}
\definecolor{trunk}{rgb}{0.5294117647058824, 0.23529411764705882, 0.0}
\definecolor{terrain}{rgb}{0.5882352941176471, 0.9411764705882353, 0.3137254901960784}
\definecolor{pole}{rgb}{1.0, 0.9411764705882353, 0.5882352941176471}
\definecolor{trafficsign}{rgb}{1.0, 0.0, 0.0}

\newcommand\semcolor[1][black]{\textcolor{#1}{\rule{2.2mm}{2.2mm}}}

\title{\LARGE \bf
Scan-based Semantic Segmentation of LiDAR Point Clouds:\\
An Experimental Study
}

\author{Larissa T. Triess$^{1,2}$, David Peter$^{1}$, Christoph B. Rist$^{1}$, and J. Marius Z\"ollner$^{2,3}$
\thanks{$^{1}$Mercedes-Benz AG, Research and Development, Stuttgart, Germany}%
\thanks{$^{2}$Karlsruhe Institute of Technology, Karlsruhe, Germany}%
\thanks{$^{3}$Research Center for Information Technology, Karlsruhe, Germany}
\thanks{Primary contact: \texttt{larissa.triess@daimler.com}}
}

\begin{document}

\maketitle
\IEEEpubidadjcol
\pagestyle{empty}

\begin{abstract}
Autonomous vehicles need to have a semantic understanding of the three-dimensional world around them in order to reason about their environment.
State of the art methods use deep neural networks to predict semantic classes for each point in a LiDAR scan.
A powerful and efficient way to process LiDAR measurements is to use two-dimensional, image-like projections.
In this work, we perform a comprehensive experimental study of image-based semantic segmentation architectures for LiDAR point clouds.
We demonstrate various techniques to boost the performance and to improve runtime as well as memory constraints.

First, we examine the effect of network size and suggest that much
faster inference times can be achieved at a very low cost to accuracy.
Next, we introduce an improved point cloud projection technique that does not
suffer from systematic occlusions. We use a cyclic padding
mechanism that provides context at the horizontal field-of-view boundaries.
In a third part, we perform experiments with a soft Dice loss function
that directly optimizes for the intersection-over-union metric.
Finally, we propose a new kind of convolution layer with a reduced amount
of weight-sharing along one of the two spatial dimensions, addressing the large
difference in appearance along the vertical axis of a LiDAR scan.

We propose a final set of the above methods with which the model achieves an increase of 3.2\% in mIoU segmentation performance over the baseline while requiring only 42\% of the original inference time.
The code can be found on our project page \textit{\href{http://ltriess.github.io/scan-semseg}{http://ltriess.github.io/scan-semseg}}.

\end{abstract}

\section{Introduction}
\label{sec:introduction}

\begin{figure}[t]
	\centering
	\includegraphics[scale=1]{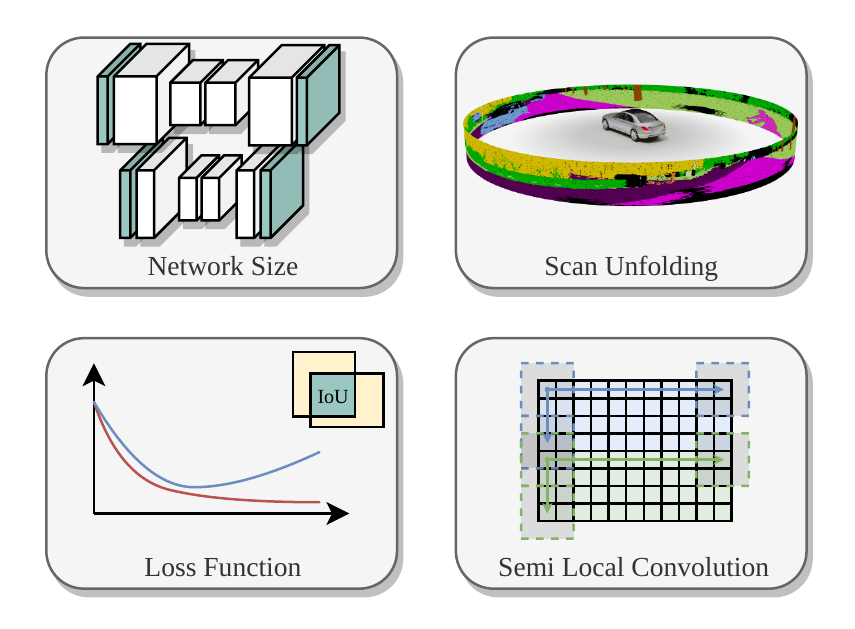}
	\caption{
	\textbf{Paper outline}:
	The experimental study is structured into four major parts. Section~\ref{sec:experiments_params} investigates effects of the network size on accuracy and runtime. Section~\ref{sec:experiments_projection} introduces an improved projection technique for LiDAR point clouds. Section~\ref{sec:experiments_loss} compares cross-entropy to soft Dice loss and section~\ref{sec:experiments_semilocal} studies the proposed SLC layer.
	}
	\label{fig:overview}
\end{figure}

\begin{figure*}[t]
    \centering
    \begin{subfigure}{0.48\linewidth}
        \centering
        \includegraphics[width=.85\linewidth]{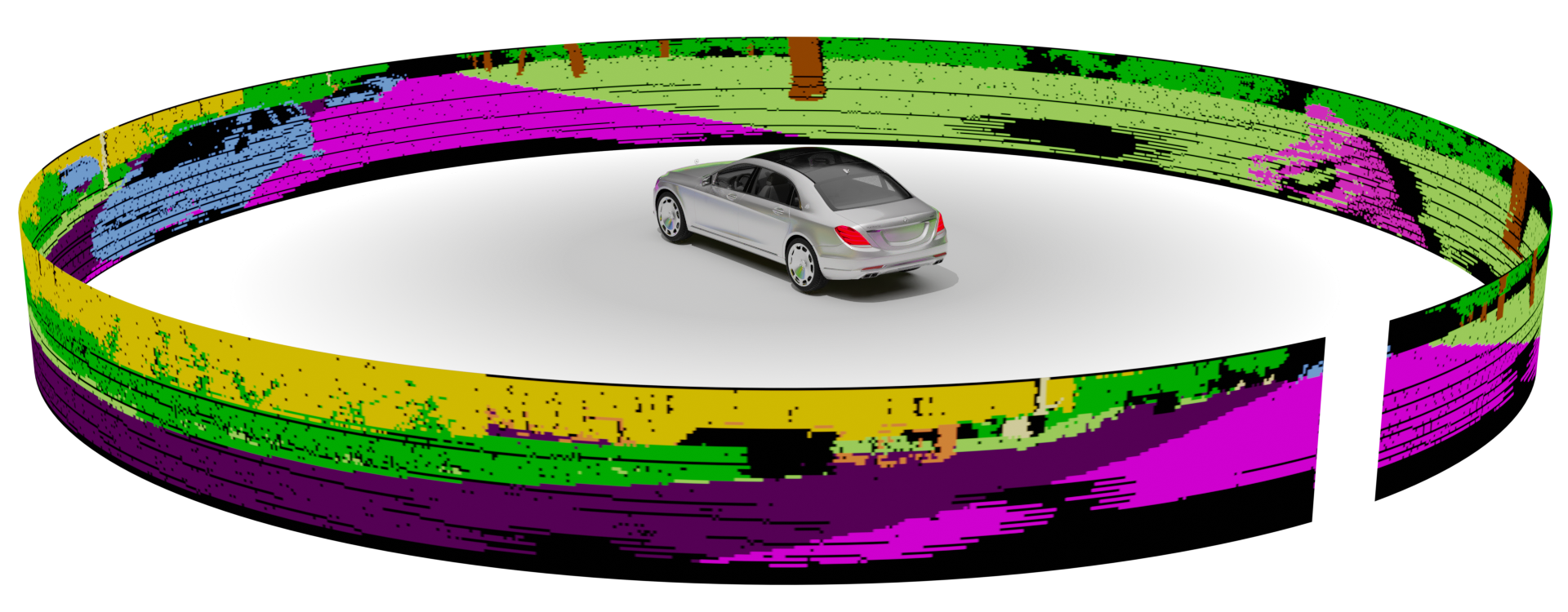}

		\vspace{0.3cm}

        \includegraphics[width=.98\linewidth]{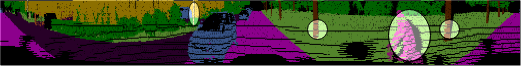}
        \caption{
        \label{fig:projection_ego}
        Ego-motion corrected projection method~\cite{milioto2019iros}
        }
    \end{subfigure}%
    \hspace{0.3cm}
    \begin{subfigure}{0.48\linewidth}
        \centering
		\includegraphics[width=.85\linewidth]{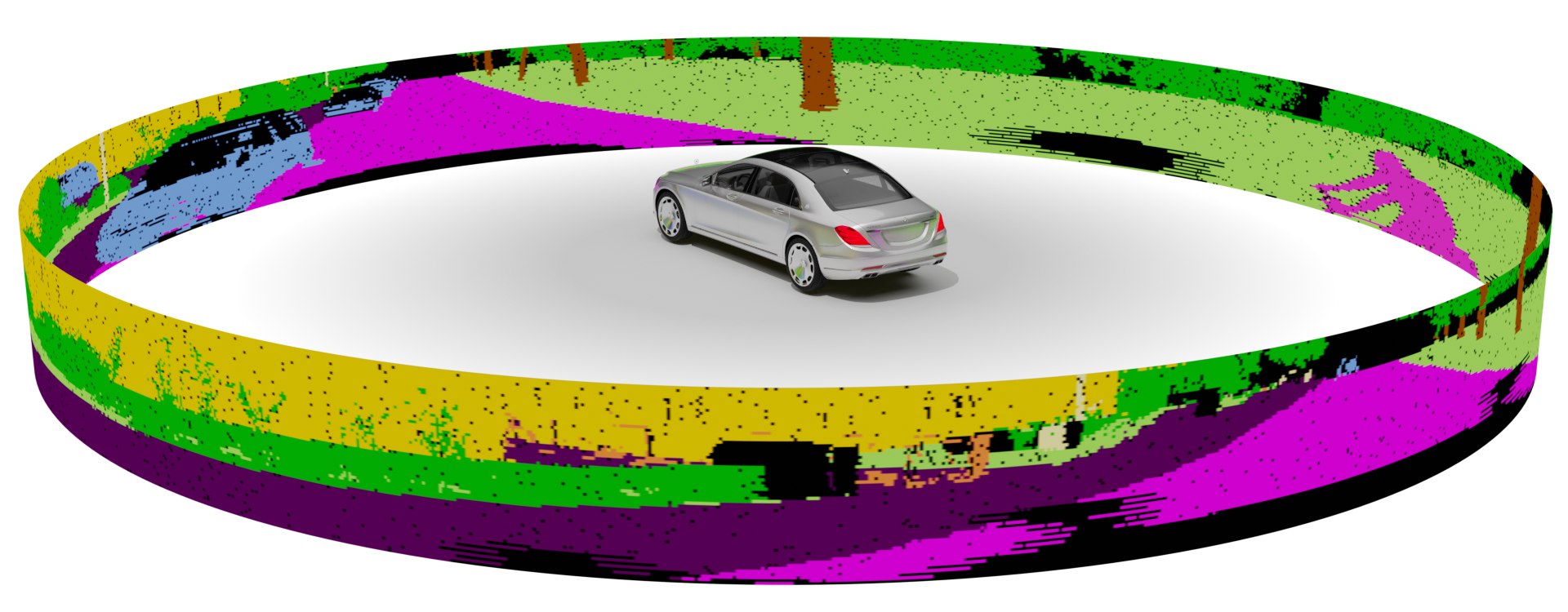}

		\vspace{0.3cm}

        \includegraphics[width=.98\linewidth]{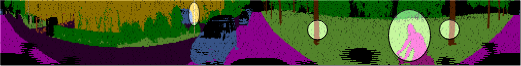}
        \caption{
        \label{fig:projection_scan}
        Scan unfolding method [Ours]
        }
    \end{subfigure}%

	\vspace{0.4cm}

    \resizebox{\linewidth}{!}{%
		\begin{tabular}{llllllllll}
			\semcolor[car] car & \semcolor[bicycle] bicycle & \semcolor[motorcycle] motorcycle & \semcolor[truck] truck & \semcolor[terrain] terrain & \semcolor[person] person & \semcolor[bicyclist] bicyclist & \semcolor[motorcyclist] motorcyclist & \semcolor[road] road & \semcolor[trafficsign] traffic-sign \\
			\semcolor[vegetation] vegetation & \semcolor[trunk] trunk & \semcolor[othervehicle] other-vehicle & \semcolor[pole] pole & \semcolor[fence] fence & \semcolor[building] building & \semcolor[otherground] other-ground & \semcolor[sidewalk] sidewalk & \semcolor[parking] parking & \semcolor[unlabeled] unlabeled \\
		\end{tabular}
	}

	\vspace{0.2cm}

    \caption{
        \textbf{Cylindrical point cloud projection}:
        (\subref{fig:projection_ego})~Correcting for ego-motion leads to a projection that suffers from systematic point occlusions as some 3D points are projected into occupied pixels (see highlighted regions in the lower row).
        Hidden points can not provide any information to the network and may not be accurately classified.
        (\subref{fig:projection_scan})~The proposed scan unfolding method provides a dense projection without systematic discretization artifacts.
        Our cyclic padding mechanism provides context at the horizontal field-of-view boundaries by closing the gap in the cylindrical projection (top right).
    }
    \label{fig:projection}
\end{figure*}

Understanding the environment perceived by a set of sensors is an essential part of all robotics applications.
For autonomous vehicles, it is important to retrieve not just the geometric shape but also the semantic
meaning of the world around them.
A complete scene understanding allows the autonomous vehicle to reason about properties of its surrounding such as the distinction between drivable and non-drivable surfaces.
In contrast to cameras that provide a flat view of the environment, LiDAR sensors directly provide a precise sampling of the three-dimensional world, without relying on daylight illumination.

State-of-the-art semantic segmentation approaches make use of traditional two-dimensional CNNs by projecting the point clouds into an image-like structure~\cite{behley2019iccv, wu2017squeezeseg, piewak2018eccv}.
This structure imitates the internal raw data representation that is used in common LiDAR sensors and which could directly be used as input to the network.
However, datasets, such as KITTI~\cite{geiger2012cvpr} or NuScenes~\cite{nuscenes2019} provide only the list of sensor measurements without indexing to the original raw format.
This requires a proxy back-projection into the image-like structure for which no unified procedure exits.
This paper proposes a scan unfolding method for KITTI that features less projection artifacts than those currently used in literature, see top right of figure~\ref{fig:overview}.
Further, the scan unfolding allows for the application of a periodic padding scheme that provides context at the horizontal field-of-view boundaries and can be propagated through the entire network.

In this study we show that the spatial stationary assumption of convolutions is still applicable to inputs with varying statistical properties over parts of the data, such as projected LiDAR scans.
These data structures exhibit similar features as aligned images for which locally connected layers have been introduced~\cite{taigman2014cvpr}.
With the introduction of Semi Local Convolutions (SLC), we show that weight sharing convolutions stay the most powerful tool for semantic segmentation, as of today.

Projection-based approaches outperform current models that operate on the raw three-dimensional data~\cite{milioto2019iros}.
In order to surpass the current baseline of a specific metric, the networks tend to become bigger in terms of more free parameters.
This can results in a declined generalization capacity, since the network partially rather "remembers" than "learns".
Further, the architectures require more resources in terms of memory and runtime in both training and inference.
Especially for autonomous robots it is vital that the components match specific resource constrains and are operable in real-time.
We show that at the expanse of very little accuracy, the resource requirements of the models can be heavily decreased.

\section{Related Work}
\label{sec:relatedwork}

Development of semantic segmentation methods for images massively increased in recent years due to the advent of deep learning.
With a rising demand for LiDAR sensors for a precise geometric inspection of the world, three-dimensional scene understanding became another major part in this field of research.
In section~\ref{sec:relatedwork_representation}, we point out various ways to represent 3D data for further processing.
Section~\ref{sec:relatedwork_semseg} introduces several methods for semantic segmentation.
A brief overview of existing convolution layers and loss functions is given in sections~\ref{sec:relatedwork_convolution} and~\ref{sec:relatedwork_losses} for later reference.

\subsection{Data Representation}
\label{sec:relatedwork_representation}

As of today, no single representation method for 3D point-clouds has prevailed.
The networks used for point-wise semantic segmentation can be divided into two categories:
(1)~projection-based networks including multi-view~\cite{lawin2017, boulch2017}, spherical~\cite{wu2017squeezeseg, wu2018squeezesegv2, milioto2019iros}, and volumetric~\cite{meng2018vvnet, rethage2019eccv, graham2019cvpr} representations, and
(2)~point-based networks including point-wise MLPs~\cite{qi2017pointnet, qi2017pointnet2, zhao2019pointweb}, convolution-based~\cite{hua1018cvpr, thomas2019kpconv, wang2018cvpr}, and graph-based~\cite{landrieu2018cvpr, wang2019cvpr} networks.
More details on point cloud representation and related architectures for 3D data are given in a survey by Guo et al.~\cite{guo2019survey}.

Behley et al.\ showed that projection-based networks outperform state-of-the-art point-based networks for point-wise semantic segmentation on LiDAR point clouds~\cite{behley2019iccv}.
In this work we focus on spherical projection-based representations and introduce a scan unfolding method applicable to KITTI data~\cite{geiger2012cvpr}.

\subsection{Semantic Segmentation}
\label{sec:relatedwork_semseg}

Semantic segmentation is a crucial part of detailed scene understanding.
Fully convolutional neural networks (FCNs) marked the breakthrough for RGB image segmentation in deep learning research~\cite{shelhamer2017tpami}.
Introduction of dilated convolutions combined with conditional random fields improved the prediction accuracy~\cite{chen2017arxiv, yu2016iclr, krahenbuhl2011nips}.
Gains on speed were mainly achieved with encoder-decoder architectures that fuse feature maps of higher layers with spatial information from lower layers or approaches that combine image features from multiple refined paths~\cite{badrinarayanan2017tpami, lin2019tpami}.

For point-wise segmentation of 3D data, many approaches evolved from their 2D ancestors by using projection-based intermediate representations of the data.
However, crucial modifications to the respective network architectures had to be introduced to fit the needs of projected data~\cite{wu2017squeezeseg, piewak2018eccv}.
Only since the recent release of SemanticKITTI~\cite{behley2019iccv}, a large scale dataset of real-world driving scenarios with point-wise semantic annotations of LiDAR scans is publicly available to facilitate the development of point-wise semantic segmentation algorithms.

\subsection{Weight Sharing in Convolution Layers}
\label{sec:relatedwork_convolution}

Convolution layers apply a filter bank on their input.
The filter weights are shared over all spatial dimensions, meaning that for every location in the feature map the same set of filters are learned.
The re-usability of weights causes a significant reduction in the number of parameters compared to fully connected layers. This allows deep convolutional neural networks to be trained successfully, in turn leading to a substantial performance boost in many computer vision applications.
The underlying premise of convolutional methods is that of translational symmetry, i.e.\ that features that have been learned in one region of the image are useful in other regions as well.

For applications such as face recognition which deal with aligned data, locally connected layers have proved to be advantageous~\cite{gregor2010, huang2012cvpr, taigman2014cvpr}.
These layers also apply a filter bank.
Contrary to convolutional layers, weights are not shared among the different locations in the feature map, allowing different sets of filters to be learned for every location in the input.

The spatial stationary assumption of convolutions does not hold for aligned images due to different local statistics in distant regions of the image.
In a projected LiDAR scan, the argument holds true for sensors that are mounted horizontally.
Each horizontal layer is fixed at a certain vertical angle.
As the environment of the sensor is not invariant against rotations around this axis, this leads to different distance statistics in each vertical layer.
To the best of our knowledge, applying locally connected filters on point cloud projections has not been investigated yet.

\subsection{Loss Functions}
\label{sec:relatedwork_losses}

For semantic segmentation tasks, the multi class cross-entropy
\begin{equation}
    CE(\hat{y}, y) = -\sum_{i,c} \hat{y}^i_c \log y^i_c
\end{equation}
is the most-often used loss function~\cite{good1956}. Here, $\hat{y}^i_c$ is the one-hot encoded ground truth distribution for class $c$ at pixel position $i$, while $y^i_c$ is the corresponding softmax prediction.

The performance of such systems is usually evaluated with the Jaccard Index over all classes~\cite{jaccard1901}, which is often referred to as mean intersection-over-union (mIoU).
In order to reach high mIoU values, the cross-entropy is minimized over training.
However, the loss does not directly reflect the inverse of the metric.

In order to directly maximize the mIoU, it is possible to use the Dice coefficient~\cite{sorensen1948, dice1945}.
The soft Dice loss can be written as
\begin{equation}
    DL(\hat{y},y) = 1 - \frac{1}{C} \sum_c \frac{2 \sum_i \hat{y}^i_c y^i_c}{\sum_i (\hat{y}^i_c)^2 + \sum_i (y^i_c)^2}
\end{equation}
where $C$ is the total number of classes.

\subsection{Contribution}
\label{sec:relatedwork_contribution}

Our main contributions are:
\begin{itemize}
\item a comprehensive study on training techniques for real-world image-based semantic segmentation architectures
\item a proposal for dense scan unfolding on KITTI and a cyclic padding mechanism for horizontal field-of-view context
\item introduction of Semi Local Convolutions, a layer with weight-sharing along only one of the two spatial dimensions
\end{itemize}

\section{Method}
\label{sec:method}

The four major components of this paper are depicted in figure~\ref{fig:overview}.
In the following both the scan unfolding method and the Semi Local Convolution are explained.

\begin{figure}[t]
    \centering
    \begin{subfigure}[b]{0.3\linewidth}
        \centering
        \includegraphics[width=\linewidth]{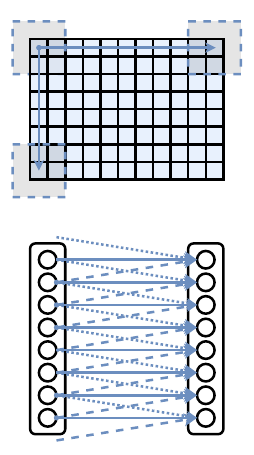}
        \caption{\label{fig:conv_alpha1}$\alpha = 1$}
    \end{subfigure}%
    \begin{subfigure}[b]{0.3\linewidth}
        \centering
        \includegraphics[width=\linewidth]{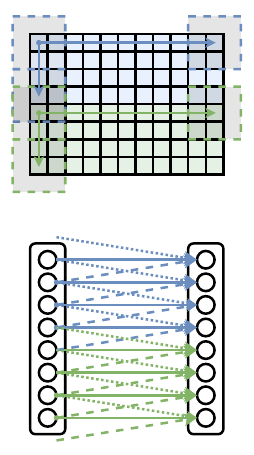}
        \caption{\label{fig:conv_alpha2}$\alpha = 2$}
    \end{subfigure}%
    \begin{subfigure}[b]{0.3\linewidth}
        \centering
        \includegraphics[width=\linewidth]{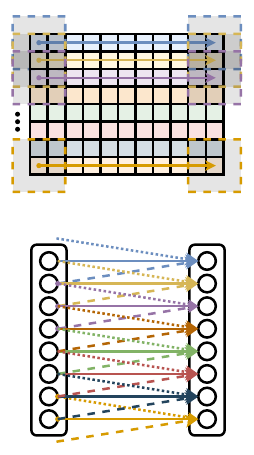}
        \caption{\label{fig:conv_alpha8}$\alpha = H_{x} = 8$}
    \end{subfigure}
    \caption{
        \textbf{Semi Local Convolution (SLC)}:
        (\subref{fig:conv_alpha1})~Illustration of a normal convolution for an input tensor of shape $[H_{x}, W_{x}] = [8, 11]$. A $3 \times 3$ sliding kernel is represented by the gray rectangle. Weight sharing is active across the full image. This
        is a special case of SLC with $\alpha = 1$.
        (\subref{fig:conv_alpha2})~SLC with $\alpha = 2$. Weights are
        shared in the upper and lower half of the input, respectively. This allows the network to learn different kernels depending on the horizontal position in the input image. (\subref{fig:conv_alpha8})~For $\alpha = H_{x}$, weight sharing along the vertical dimension is completely turned off, weights are only shared horizontally. Different filters can be learned for each individual vertical position.
    }
    \label{fig:conv}
\end{figure}

\subsection{Scan Unfolding}
\label{sec:method_projection}

The LiDAR sensor considered in this work consists of vertically stacked send-and-receive modules which revolve around a common vertical axis.
While rotating, each module periodically measures the distance and reflectivity at its current orientation.
Internally, the sensor represents the raw data in a range-image-like fashion.
Openly available datasets provide the data as lists of Cartesian coordinates~\cite{geiger2012cvpr, nuscenes2019}.
This requires a back-projection into the image-like structure for projection-based networks.
Figure~\ref{fig:projection} shows two different projection schemes: ego-corrected projection and our scan unfolding.

\subsubsection{Ego-motion corrected Projection}
\label{sec:method_projection_ego}
The projection shown in figure~\ref{fig:projection_ego} is a proxy representation by Milioto et al.~\cite{milioto2019iros}.
It suffers from mutual point occlusions due to the ego-motion correction of the data and leaves large areas without data (black pixels).

\subsubsection{Scan Unfolding}
\label{sec:method_projection_scan}
Figure~\ref{fig:projection_scan} depicts a projection with reduced mutual point occlusions, thus minimizing the loss of information.
The scan unfolding method is designed to be a proxy representation of the original raw sensor data.
The conducted back-projection is only necessary since the dataset does not provide the direct sensor output or an index-map for simple back-projection.
When working in an actual autonomous driving stack, the preprocessing needed for the scan unfolding can be omitted, as the LiDAR scanner directly provides the depth-image format.
We provide algorithm~\ref{algo:projection} that exploits the distinct data representation of the KITTI dataset to generate the desired scan pattern.
The algorithm is applied to the uncorrected scan data (without ego-motion compensation), which is accessible via the raw data of KITTI\footnote{Sequence 3 of the Odometry Benchmark used for SemanticKITTI is not published in the raw data, thus all experiments conducted in this paper omit sequence 3 in training.}.
The KITTI raw format lists LiDAR points of an accumulated 360 degree scan in order of their vertical index of the associated sensor scan line.
However, the crossovers between two consecutive scan lines happen at the cut to the rear of the vehicle and are not indicated in the provided data.
Thus the task of detecting these positions to assign each point to its vertical index remains and is addressed by our algorithm.

\begin{algorithm}[t]
    \SetAlgoLined
    \DontPrintSemicolon

    \SetKwProg{Fn}{Function}{\string:}{}
    \SetKwFunction{GetRows}{GetRows}
    \SetKwFunction{GetColumns}{GetColumns}

    \KwData{An array \textit{points} of size $N\times 3$, a tuple $(H, W)$}
    \KwResult{\textit{projection} of \textit{points} with shape $H\times W$}
    \BlankLine

    \textit{depth} $\longleftarrow \sqrt{\textit{points}_x^2 + \textit{points}_y^2 + \textit{points}_z^2}$\;
    \textit{rows} $\longleftarrow$ \GetRows{\textit{points}}\;
    \textit{columns} $\longleftarrow$ \GetColumns{\textit{points}}\;
    sort $\textit{columns}$, $\textit{rows}$ and $\textit{depth}$ by decreasing \textit{depth}\;
    \textit{projection} $\longleftarrow$ array of shape $H\times W$\;
    \textit{projection}[$\textit{columns}$, $\textit{rows}$] = \textit{depth}\;

    \BlankLine
    \Fn(){\GetRows{points}}{
        $\phi \longleftarrow \text{atan2}(\textit{points}_y,\,\textit{points}_x)$\;
        \textit{jump} $\longleftarrow \left|\phi[1:] - \phi[:-1]\right| > \textit{threshold}$\;
        \textit{jump} $\longleftarrow$ [0] + \textit{jump}\;
        \textit{rows} $\longleftarrow$ cumulative sum over \textit{jump}\;
        \Return{rows}\;
    }

    \BlankLine
    \Fn(){\GetColumns{points}}{
        $\phi \longleftarrow \text{atan2}(\textit{points}_y,\,\textit{points}_x)$\;
        \textit{columns} $\longleftarrow W\cdot (\pi - \phi) / (2\pi)$\;
        \Return{columns}\;
    }
    \caption{\textbf{Scan Unfolding on KITTI}: \textit{threshold} is chosen to be larger than the horizontal resolution (KITTI: $threshold = 0.3\degree$).}
    \label{algo:projection}
\end{algorithm}

When convolving over the data the input is usually padded in order to match the desired output shape.
Since LiDAR measurements represent a constant stream of data along the horizontal axis of the projections, the precise padding would take snippets from the previous and subsequent $360\degree$ scan in time.
This is not practical when training the network and not applicable at inference time.
However, using the scan-based projection we can implement a cyclic padding strategy by basically taking the values from the opposite side of the range image.
Due to the cylindrical projection of the scan, a closed $360\degree$ view is formed (see figure~\ref{fig:projection_scan}).
This can be propagated through the entire network.

\subsection{Semi Local Convolution (SLC)}
\label{sec:method_slc}

In order to introduce SLCs, consider an input feature map $x$ with shape $\left[H_{x}, W_{x}, C_{x}\right]$, representing a cylindrical projection with height $H_x$, width $W_x$ and $C_{x}$ channels.
The output of the layer is another feature map~$y$ with shape $\left[H_{y}, W_{y}, C_{y}\right]$. In the following, without loss of generality, we consider $x$ to be padded such that $H_{y} = H_{x}$ and $W_{y} = W_{x}$.

In a normal convolution layer with a kernel $k$ of shape $\left[I, J, C_x, C_{y}\right]$, the output would be
\begin{equation}
\label{eq:conv}
    y_{h, w, c_{y}} = \sum_{c_x}\sum_{i}\sum_{j} k_{i, j, c_x, c_{y}} \cdot x_{h-i, w-j, c_{x}}
\end{equation}
where the sum over $i$ (and similarly for $j$) is appropriately restricted to the range $-\lfloor I/2 \rfloor \dots \lfloor I/2 \rfloor$.

In a SLC layer, the kernel has multiple components for different parts along the vertical axis of the input as illustrated in figure~\ref{fig:conv} (note that the concept can also be applied to the horizontal direction).

With $\alpha \in \mathbb{N}$ the number of components ($1 \leq \alpha \leq H_{x}$), the kernel has a shape of $\left[I, J, C_x, C_{y}, \alpha\right]$. The output of the SLC is then given by
\begin{equation}
\label{eq:semi-conv}
    y_{h, w, c_{y}} = \sum_{c_{x}}\sum_{i}\sum_{j} k_{i, j, c_x, c_{y}, \alpha_h} \cdot x_{h-i, w-j, c_{x}}
\end{equation}
where $\alpha_h = \lfloor h/H \cdot \alpha \rfloor$ selects the respective filter-component depending on the vertical position $h$.

For $\alpha = H$, there is no weight sharing along the vertical axis, a new filter is used for every single data row.
For $\alpha = 1$, we obtain a regular convolution as defined in equation~\ref{eq:conv}.
For values in between, the degree of weight sharing can be adapted to the desired application.

\section{Experiments}
\label{sec:experiments}

\begin{table*}
	\caption{
		\textbf{Semantic segmentation performance:}
		This table shows experimental results for a subset of the proposed techniques and compares them with RangeNet$^\star$ (R$^\star$).
		Note that the numbers deviate from the ones published in~\cite{milioto2019iros}, as we report on the validation dataset instead of the test dataset.\\
        \footnotesize{$^\star$ we drop $x$, $y$, and $z$ channels from the input as our experiments showed that these features do not influence the performance in a significant way.}
	}
	\centering
	\setlength{\tabcolsep}{4pt}
	\resizebox{\textwidth}{!}{
		\begin{tabular}{lccc | cc | ccccccccccccccccccc}
			\toprule
				\rotatebox{90}{base network}
				& \rotatebox{90}{Dice loss}
				& \rotatebox{90}{scan unfolding}
				& \rotatebox{90}{cyclic padding}
				& \rotatebox{90}{\textbf{inference} $\left[\frac{\text{ms}}{\text{frame}}\right]$}
				& \rotatebox{90}{\textbf{mean IoU} [\%]}
				& \rotatebox{90}{\semcolor[bicycle] bicycle}
				& \rotatebox{90}{\semcolor[car] car}
				& \rotatebox{90}{\semcolor[motorcycle] motorcycle}
				& \rotatebox{90}{\semcolor[truck] truck}
				& \rotatebox{90}{\semcolor[othervehicle] other-vehicle}
				& \rotatebox{90}{\semcolor[person] person}
				& \rotatebox{90}{\semcolor[bicyclist] bicyclist}
				& \rotatebox{90}{\semcolor[motorcyclist] motorcyclist}
				& \rotatebox{90}{\semcolor[parking] parking}
				& \rotatebox{90}{\semcolor[road] road}
				& \rotatebox{90}{\semcolor[sidewalk] sidewalk}
				& \rotatebox{90}{\semcolor[otherground] other-ground}
				& \rotatebox{90}{\semcolor[building] building}
				& \rotatebox{90}{\semcolor[pole] pole}
				& \rotatebox{90}{\semcolor[trafficsign] traffic-sign}
				& \rotatebox{90}{\semcolor[fence] fence}
				& \rotatebox{90}{\semcolor[trunk] trunk}
				& \rotatebox{90}{\semcolor[terrain] terrain}
				& \rotatebox{90}{\semcolor[vegetation] vegetation} \\
			\midrule
				R$^\star$ & & &
				& 74.3 & 46.7
				& 23.0 & 91.0 & 31.8 & 29.5 & 29.6
				& 26.2 & 48.4 & 0.0
				& 41.5 & 92.9 & 78.9 & 0.4
				& 82.1 & 36.1 & 25.7
				& 49.7 & 42.9 & 75.5 & 82.7  \\
			\specialrule{0.5pt}{1pt}{2pt}
				R$^\star$ & \checkmark & &
				& 74.3 & 48.2
				& 24.3 & 92.0 & 28.1 & 39.5 & 25.6
				& 17.5 & 55.6 & 0.0
				& 36.9 & 92.4 & 78.5 & 0.0
				& 81.9 & 47.2 & 34.6
				& 48.3 & 53.5 & 75.0 & 84.0  \\
				R$^\star$ & & \checkmark &
				& 74.3 & 47.5
				& 23.9 & 90.7 & 37.6 & 31.3 & 24.9
				& 22.9 & 53.0 & 0.0
				& 43.2 & 93.2 & 79.2 & 0.3
				& 83.5 & 36.2 & 25.8
				& 51.2 & 45.9 & 75.4 & 84.0  \\
				R$^\star$ & & \checkmark & \checkmark
				& 74.3 & 47.9
				& 23.1 & 92.1 & 32.3 & 35.5 & 22.8
				& 24.9 & 51.5 & 0.0
				& 43.0 & 94.8 & 79.9 & 0.3
				& 84.2 & 36.3 & 25.4
				& 49.4 & 47.9 & 77.2 & 84.1  \\  
				R$^\star$ & \checkmark & \checkmark & \checkmark
				& 74.3 & \textbf{48.5}
				& 22.1 & 93.3 & 26.0 & 29.3 & 21.9
				& 15.3 & 41.8 & 0.2
				& 38.1 & 93.1 & 77.7 & 0.7
				& 82.1 & 45.8 & 38.2
				& 50.1 & 49.9 & 74.3 & 84.2  \\  
			\specialrule{0.5pt}{1pt}{2pt}
				D & \checkmark & \checkmark & \checkmark
				& \textbf{30.9} & 48.2
				& 25.5 & 91.1 & 25.6 & 38.8 & 21.7
				& 23.0 & 48.6 & 0.0
				& 43.3 & 93.1 & 77.9 & 0.5
				& 82.9 & 48.6 & 37.3
				& 55.9 & 48.3 & 70.8 & 83.3  \\
			\bottomrule
            \label{tab:experiments_overall}
		\end{tabular}%
	}
\end{table*}

This section is structured into five parts, four of which represent the components illustrated in figure~\ref{fig:overview}.
In the end, we give a brief summary on the experiments and introduce a model that incorporates the positive findings of this study.

The basis of our experiments is the RangeNet implementation of Milioto et al.~\cite{milioto2019iros}.
Note that we compare our models against a slightly modified version of RangeNet, referred to as RangeNet$^\star$ (R$^\star$), which omits $x$, $y$, and $z$ as input channels.
We benchmarked both against each other and found no significant difference in the resulting metric results.
The first row of table~\ref{tab:experiments_overall} shows the baseline results of RangeNet$^\star$.

\subsection{Network Parameters}
\label{sec:experiments_params}

Larger networks tend to be more prone to overfitting.
RangeNet with its 50.4 million trainable parameters is also affected by this.
Figure~\ref{fig:experiments_paramsiou} and table~\ref{tab:experiments_paramsiou} show the performance of the network for a decreasing number of trainable parameters by adapting the filter sizes within the convolutions (details are given in the appendix).
A large reduction of parameters, causes the performance to decrease only slightly.
Further, we observe that the smaller networks generalize better due to decreased overfitting.
With a reduction to only 10\% of the original number of parameters, we still reach 96\% of the performance while, at the same time, decreasing the inference time of the network to one third.

\begin{figure}
    \centering
    
    \captionof{table}{
    \textbf{Performance for different network sizes}:
    We report the mean value of training and validation mIoU as well as the respective standard deviation $\left(\pm x\right)$.
    The network configurations are given in the appendix.
    }
    \resizebox{\linewidth}{!}{ 
    \begin{tabular}{l ccccc}
        \toprule
			Number of		 & A & B & C & D & RangeNet$^\star$ \\
			parameters		 & 0.4M & 1.3M & 4.2M & 12.7M & 50.4M \\
        \midrule
            Train mIoU [\%] & 39.2 $\pm$ 0.5 & 45.6 $\pm$ 1.0 & 52.0 $\pm$ 1.3 & 54.1 $\pm$ 3.2 & 59.7 $\pm$ 4.1 \\
            Val mIoU [\%]   & 38.7 $\pm$ 0.6 & 41.7 $\pm$ 5.1 & 43.5 $\pm$ 2.5 & 44.7 $\pm$ 1.2 & 46.4 $\pm$ 0.7 \\
            Inference time [ms] & 20.5 & 22.1 & 23.9 & 30.9 & 74.3 \\
        \bottomrule
    \end{tabular}
    } 
    \label{tab:experiments_paramsiou}
    
    \vspace{0.4cm}

    \begin{tikzpicture}
\pgfplotsset{
	width=7.4cm,
	height=4.6cm,
	compat=newest,
	grid style={dashed,gray!30},
}

\definecolor{inftime}{RGB}{0,153,115}
\definecolor{train}{RGB}{184,84,80}
\definecolor{val}{RGB}{108,142,191}

\begin{semilogxaxis}[
	legend cell align=left,
	axis y line*=left,
	grid=both,
	ymin=0,
	ymax=70,
	xmin=1e5,
	xmax=1e8,
	xlabel=Number of parameters,
	ylabel=mIoU,
	y unit=\%,
	legend style={
		at={(0,1.05)},
		anchor=south,
		legend columns=-1},
]

\addplot+[only marks, color=train, mark options={train}, error bars/.cd, y dir=both, y explicit]
table[y=train_miou, x=num_params, y error=train_stddev, col sep=comma] {data/params_iou.csv};

\addplot+[only marks, color=val, mark options={val}, error bars/.cd, y dir=both, y explicit]
table[y=val_miou, x=num_params, y error=val_stddev, col sep=comma] {data/params_iou.csv};

\legend{train, val}

\end{semilogxaxis}

\begin{semilogxaxis}[
axis y line*=right,
	ymin=0,
	ymax=140,
	xmin=1e5,
	xmax=1e8,
	ylabel=Time per frame,
	y unit=s,
	y SI prefix=milli,
	y filter/.code={\pgfmathmultiply{#1}{1000}},
	ytick={0,40,80,120},
	xtick=\empty,
	legend style={
		at={(1,1.05)},
		anchor=south,
		legend columns=-1},
]

\addplot [only marks, mark=diamond*, color=inftime, mark options={inftime}]
table[y=inference_time_per_frame, x=num_params, col sep=comma] {data/params_iou.csv};
\label{plot_time}

\legend{inference time}

\end{semilogxaxis}

\end{tikzpicture}
    
    \captionof{figure}{
    \textbf{Overfitting}:
    A significant overfitting gap is present for networks at RangeNet size.
    The effect only vanishes when reducing the number of parameters by two orders of magnitude.
    }
    \label{fig:experiments_paramsiou}
\end{figure}

\subsection{Loss Functions}
\label{sec:experiments_loss}

The second row of table~\ref{tab:experiments_overall} shows that replacing cross-entropy loss with Dice loss increases the mean IoU by 3.2\%.
Class-wise the two losses show distinguished quality.
Dice loss reaches better performances on classes bicycle, bicyclist, pole, traffic-sign, and trunk.
Cross-entropy, on the other hand, performs better on motorcycle, parking, and person.
If IoU is the metric to reflect the desired quality in a network performance, it is advisable to use Dice loss instead of cross-entropy.
It has the advantage of directly maximizing the metric as opposed to cross-entropy.

\subsection{Scan construction}
\label{sec:experiments_projection}

We compare the ego-motion corrected projection with our scan unfolding method in two otherwise identical settings.
The former uses the ego-motion corrected data from SemanticKITTI, while the latter uses the raw data obtained from KITTI.
The point-wise annotations are identical for both.
However, note the target segmentation might differ depending on the occlusions that arise from the projection.
Table~\ref{tab:experiments_overall} shows the validation results for our scan unfolding method (row three) in comparison to RangeNet$^\star$ using the ego-motion corrected data.
The scan unfolding achieves a gain of 1.7\% in mean IoU.
Classes with small or thin objects, such as bicyclist or trunk, benefit especially.
This can be attributed to the differences in projection for foreground objects, as highlighted in figure~\ref{fig:projection}.

In addition, we replace zero padding with our cyclic padding strategy in all convolution layers.
The results are listed in the fourth row of table~\ref{tab:experiments_overall}.
Exploiting the cycle consistency of the scan renders beneficial for the performance but does not generate a substantial boost.
We propose this as a more accurate padding scheme than the default zero-padding for 360\degree scans.

\subsection{Semi Local Convolutions}
\label{sec:experiments_semilocal}

We investigated the introduction of SLC layers in various experiments.
In general, we did not find evidence that SLCs can outperform normal convolutions.
The performance usually decreased with increasing $\alpha$, with a stronger effect for larger networks. We attribute this to the fact that the number of parameters in such a layer increases with $\alpha$.
Only very small networks showed an improvement when using SLCs with $\alpha = 2$ in the output head of the network.
We conclude that normal convolution layers of adequate capacity can already handle the different statistical properties across the vertical spatial dimension.
We believe that is still worth to report these results.

\subsection{Summary}
\label{sec:summary}

Considering the above insights, we combined components that generated a positive effect on the segmentation accuracy.
Table~\ref{tab:experiments_overall} shows that combining Dice loss and the scan unfolding method with cyclic padding reaches the best performance.
We further tested these settings on a smaller network D (see table~\ref{tab:experiments_paramsiou}) and achieved a higher segmentation score than with the plain version of the much larger RangeNet$^\star$.
The inference time of this model is less than half of the time of the bigger model.

\section{Conclusion}
\label{sec:conclusion}

This paper presents an experimental study on projection-based semantic segmentation of LiDAR point clouds.
Our experiments show that specially chosen loss functions and input data representations can lead to a boost in semantic segmentation performance.
We advocate our scan unfolding method over the cylindrical projection of ego-motion corrected data.
In the case of single-frame processing, it can be combined with a cyclic padding mechanism which leads to another small improvement.

We also demonstrated that the network size can be drastically reduced at very little cost to accuracy, allowing for applications on hardware with limited resources or hard real-time constraints.
By combining Dice loss and our scan unfolding method with cyclic padding, we propose a fast network architecture that outperforms much slower state-of-the-art networks without these modifications.


\section*{APPENDIX}
\begin{table}[h]
	\centering
	\caption{
    \textbf{Network configuration}:
    Filter size configuration for the encoding blocks in the backbone.
    The networks correspond to the ones in table~\ref{tab:experiments_paramsiou}.
    }
	\begin{tabular}{l rrrrr}
        \toprule
			Network			 &  A &  B &   C &   D & R$^\star$ \\
        \midrule
            Filter Sizes	 & 32 & 32 &  32 &  32 &   32 \\
            				 & 32 & 48 &  48 &  48 &   64 \\
            				 & 32 & 64 &  64 &  64 &  128 \\
            				 & 32 & 64 &  96 & 128 &  256 \\
            				 & 32 & 64 & 128 & 256 &  512 \\
            				 & 32 & 64 & 256 & 512 & 1024 \\
        \bottomrule
    \end{tabular}
\end{table}


\bibliographystyle{IEEEtran}
\bibliography{IEEEabrv,refs}

\end{document}